\documentclass[10pt,twocolumn,letterpaper]{article}

\usepackage{iccv}
\usepackage{times}
\usepackage{epsfig}
\usepackage{graphicx}
\usepackage{amsmath}
\usepackage{amssymb}

\usepackage[breaklinks=true,bookmarks=false]{hyperref}

\iccvfinalcopy 

\def\iccvPaperID{****} 

\ificcvfinal\fi

\begin{document}

\title{Unsupervised Multi-stream Highlight detection for the Game "Honor of Kings"}

\author{Li Wang, Zixun Sun, Wentao Yao, Hui Zhan, Chengwei Zhu\\
Interactive Entertainment Group, Tencent Inc.\\
{\tt\small \{willyliwang, zixunsun, wentaoyao, huizhan, chavezzhu\}@tencent.com}
}

\maketitle
\ificcvfinal\thispagestyle{empty}\fi

\begin{abstract}
   With the increasing popularity of E-sport live, Highlight Flashback has been a critical functionality of live platforms, which aggregates the overall exciting fighting scenes in a few seconds. In this paper, we introduce a novel training strategy without any additional annotation to automatically generate highlights for game video live. Considering that the existing manual edited clips contain more highlights than long game live videos, we perform pair-wise ranking constraints across clips from edited and long live videos. A multi-stream framework is also proposed to fuse spatial, temporal as well as audio features extracted from videos. To evaluate our method, we test on long game live videos with an average length of about 15 minutes. Extensive experimental results on videos demonstrate its satisfying performance on highlights generation and effectiveness by the fusion of three streams.
\end{abstract}

\section{Introduction}

The past decade has seen a rapid development of the live broadcast market, especially the e-sports market such as 'Douyu', 'Kuaishou' and 'Penguin E-sports' in China. One of the core function on live platforms is Highlight Flashback, which aims at showing the most exciting fighting clips during the lengthy live broadcast. However, the current highlight flashbacks on platforms are manually edited and uploaded. This content generation process consumes considerable human resources. Hence, automatic highlights generation is an urgent demand for these live platforms. 


\begin{figure}[t]
  \centering
  \includegraphics[width=.95\linewidth]{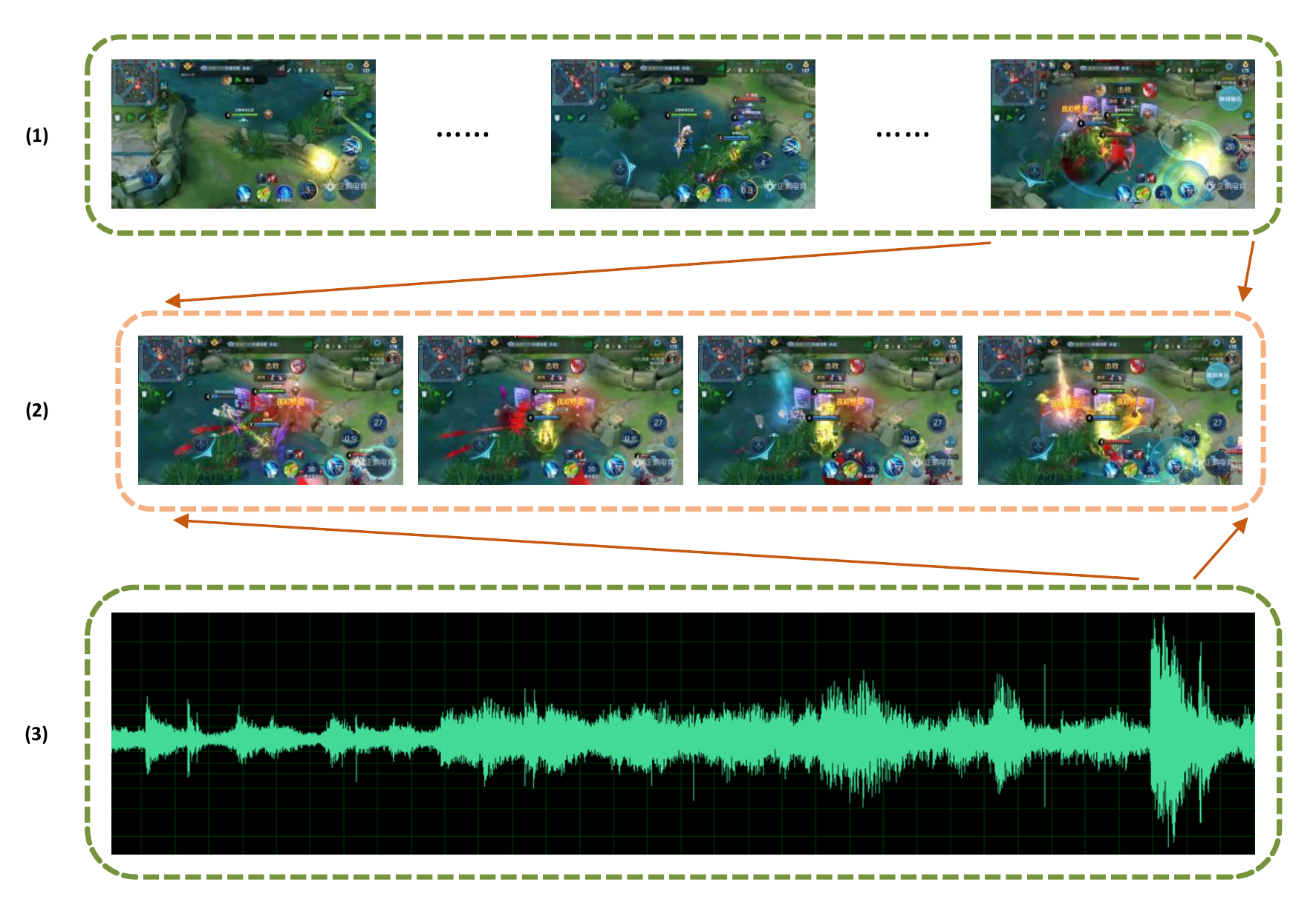}
  \caption{Examples of highlights in the live videos of the game 'Honor of Kings'. 1)The raw long live video; 2) Highlight segments corresponded to the original raw video; 3) Audio clips of the video.}
    \vspace{-0.1in}
  \label{fig:introduction}
\end{figure}

To tackle the issue above, previous works explored highlight detection from frame-level or clip-level \cite{sun2014ranking,gygli2016video2gif,yao2016highlight,fu2017video,ringer2018deep,xiong2019less}. \cite{fu2017video} addressed highlights detection task as a classification task, that was, highlight parts were regarded as target class and the rest were background. This method needed accurate annotations for each frame or clips, which was always used under supervised pattern. \cite{ringer2018deep} regarded highlight as a novel event for every frame in a video. A convolutional autoencoder was constructed for visual analysis of the game scene, face, and audio. On the other hand, \cite{gygli2016video2gif,yao2016highlight,xiong2019less} took use of the internal relationship between highlight clips and non-highlight clips, where highlight clips got higher scores than the non-highlight ones. Based on this point, a ranking net was employed to implement the relationship under both supervised and unsupervised pattern. 

In this paper, we focus on the highlight generation for the live videos of the game 'Honor of Kings'. We define the intense fight scenes in videos as highlights, at which audiences are excited. Example is demonstrated in Figure \ref{fig:introduction}. However, laborious annotations are needed for training a network specially designed for the game scenes. Thus, we take a novel training strategy for network learning using the existing videos downloaded from the 'Penguin E-sports' without any additional annotations. 

\begin{figure*}
\centering
  \includegraphics[width=.8\linewidth]{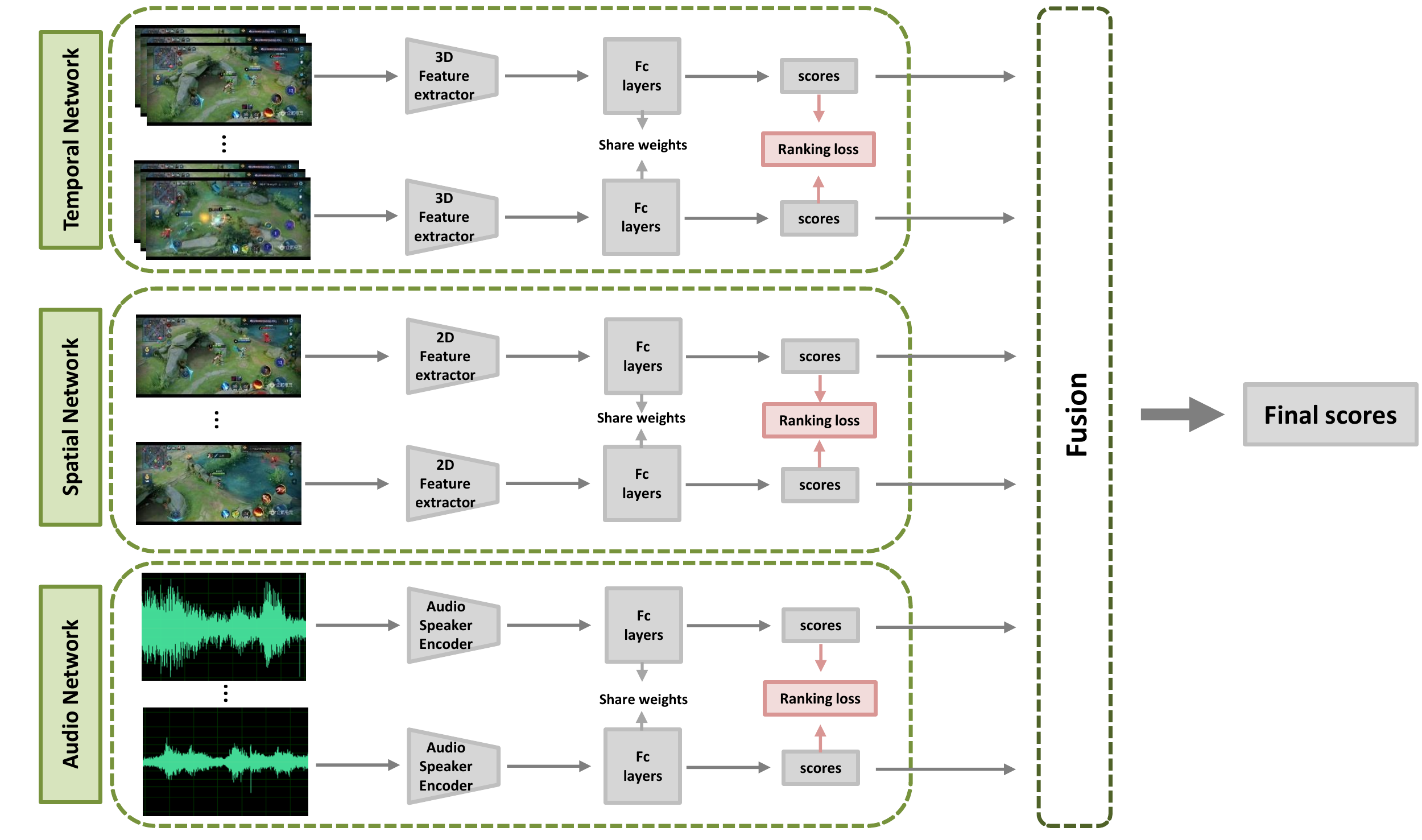}
  \caption{Overview of our framework. Three streams are trained using ranking loss, respectively. And during the inference time, we first get three scores for the target video clip and then fuse them together to obtain the final score.}
    \vspace{-0.1in}
\label{fig:network}
\end{figure*}

Considering that the unlabeled videos are hard to be classified as highlight or non-highlight, we adopt the ranking net as our basic model. Since a significant number of highlight videos are edited and uploaded to the platform by journalists and fanciers, we take clips from edited videos as the positive samples, and clips from long videos as the negative samples for our network. Since the long videos also contain high clips, which may introduce much noise for data training, we use Huber loss \cite{gygli2016video2gif} to further reduce the impact of noisy data.

Besides, as shown in Figure \ref{fig:network}, a multi-stream network is constructed to take full use of the temporal, spatial and audio information of the videos. We use simple convolutional layers to produce final highlights from temporal aspects, while fuse the auxiliary audio and spatial components to further fine-tune the results.

Our contributions can be summarized as follow:
\begin{itemize}
\item A novel training strategy for network training is proposed, which uses existing downloaded videos without any additional annotations.
\item A multi-stream network containing temporal, spatial and audio components is constructed, which takes full use of the information from the game videos.
\item Further experiment results on the game videos demonstrate the effectiveness of our method.
\end{itemize}


\section{Dataset Collection}
Existing highlight datasets \cite{sun2014ranking,song2015tvsum} contain variety of videos in natural scenes, which have great gap with the scenes in game videos. For this reason, we recollect the videos for the target scenes. We collect long raw live videos and highlight videos from Penguin E-sports platform. For network training, first, we randomly select 10 players and query their game videos, then 450 edited highlight videos and 10 long raw videos are then downloaded. The highlight videos are average 21 seconds long while the lengths of raw videos range from 6 to 8 hours. 
Due to the extreme length of the raw live videos, we randomly intercept 20 video clips from videos with an average length of 13 minutes each, so that the positive and negative samples can be balanced. Note that each video clip contains both the highlight and non-highlight intervals, where highlight clips account for about 20\% of the whole video. For testing, We download another four long videos from different players, and got one video clip from each long video with 15 minutes. Specially, the evaluated video clips contain different master heroes so that the scenarios varies dramatically and the task becomes a challenge. To evaluate the effectiveness of our approach, we annotate the evaluated video clips on the second-level. The annotated videos have 55 highlight time periods totally with average 7.83 seconds for each period.


\section{Methodology}
In this section, we introduce our multi-stream architecture which is demonstrated in Figure \ref{fig:network}.
We combine three components for highlight generation: Temporal Network extracts the temporal information; Spatial Network gets the spatial context information for each frame; Audio Network filters the unrelated scenes by utilizing the internal most powerful sound effect, which reveals the player's immersion. All of the three networks use the ranking framework, which constrains the scores produced from positive and negative samples.

\vspace{0.05in}\noindent \textbf{Temporal Network.} In this component, we exploit temporal dynamics using 3D features. We extract the features from the output of final pooling layer of ResNet-34 model \cite{he2016deep} with 3D convolution layers \cite{hara2018can}, which is pre-trained on the Kinetics dataset \cite{carreira2017quo}. The inputs for the network are video clips with 16 frames each. After the corresponding features are obtained, three fully-connected layers with the channels  \{512, 128, 1\} are added to perform the ranking constraint. 

\vspace{0.05in}\noindent \textbf{Spatial Network.} The spatial and context information plays an important role in object classification and detection tasks, which is also critical for highlight detection by providing distinguish appearances for different scenes and objects. Therefore, we set up a stream to classify highlight or non-highlight video frames. 
Different from the Temporal Network, we train the Spatial Network on frame level. A fixed-length feature is firstly extracted for each frame and then fed into the spatial ranking net. Here, we use the AlexNet \cite{krizhevsky2012imagenet}, which is pre-trained on 1.2 million images of ImageNet challenge dataset \cite{deng2009imagenet}, to generate 9216-dimension feature from the last pooling layer. And seven fully-connected layers are followed with the channels \{9216, 4096, 1024, 512, 256, 128, 64, 1\} respectively. 

\begin{figure*}[t]
\begin{center}
\includegraphics[width=.98\linewidth]{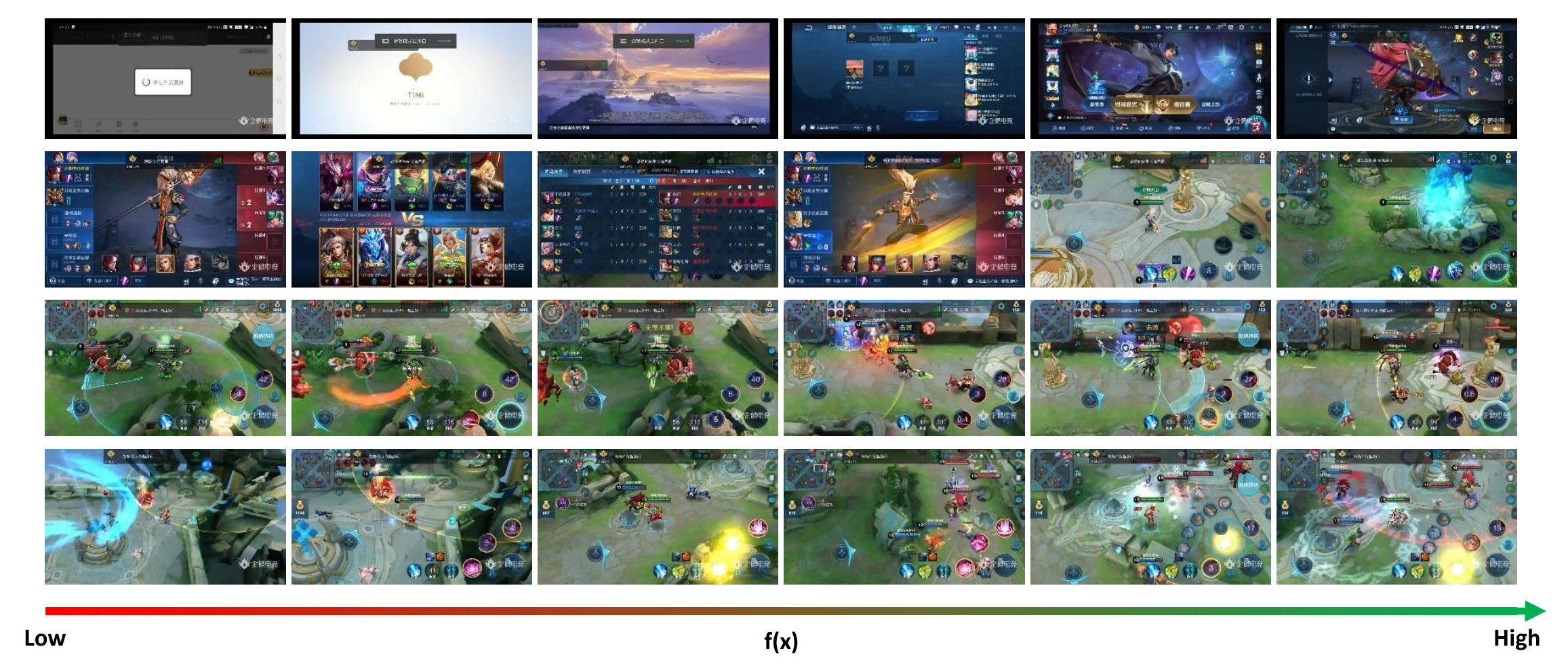}
\end{center}
   \caption{Quantitative highlight generation results. Each row demonstrates the inference scores $f(x)$ for frames from low to high in a test video. The first two rows show several non-game frames which get lower scores than the game frames. And the last two rows show the ranking scores between game frames. Obviously, the frames from intense fight scenes obtain higher scores than that from non-fight scenes.}
\label{fig:results}
\end{figure*}
 
\vspace{0.05in}\noindent \textbf{Audio Network.} It is observed that  different scenes in game live contain different characteristics. For example, the highlight clips may be immersed in the audible screams and sounds of struggle while the non-highlight parts share the more quiet property. Each one-second audio is firstly fed into a pre-trained Speaker Encoder \cite{jia2018transfer} to generate a 256-dimension feature. Then two fully-connected layers with the channels \{64, 1\} are followed.

\vspace{0.05in}\noindent \textbf{Training \& Inference process.} We train three streams separately, and samples from highlights and non-highlights satisfy the constraint:
\begin{equation}
  f(x^+) > f(x^-), \forall(x^+, x^-) \in D\\
\end{equation}

where $x^+, x^-$ are the features of input samples (frames or video clips) from highlights and non-highlights videos, respectively.  $f(*)$ denotes the ranking network, and $D$ is the Dataset.

Therefore, to optimize the networks, we employ the ranking loss between input positive and negative samples, which is described as follow:
\begin{equation}
  L_{p}(x^+, x^-) = max(0, 1-f(x^+) + f(x^-))^p, \\
\label{equation:loss}
\end{equation}

The loss function is gained under the condition that the negative samples are non-highlights. However, in our case, the input clips from live videos can also be highlights. Thus, we apply the Huber loss proposed in \cite{gygli2016video2gif} to decrease the negative effect of outliers:
\begin{equation}
   L_{Huber}(x^+, x^-) = \left\{
\begin{array}{rcl}
\frac{1}{2}L_2(x^+, x^-),       &      & \mu \leq \delta\\

\delta L_1(x^+, x^-) -  \frac{1}{2} \delta^2,    &      & otherwise
\end{array} \right.
\end{equation}

where $\mu = 1 - f(x^+) + f(x^-)$, so that the losses for outliers are small than the normal ones. $L_1$ and $L_2$ are the loss functions where $p=1$ and $p=2$ in equation (\ref{equation:loss}), respectively. Here we set $\delta = 1.5$ to distinct the outliers.

During the inference, after the scores are got from three streams, we need to fuse them to form the final scores. Here, the simple weighted summation is used with the weight \{0.7, 0.15, 0.15\} for temporal, spatial, and audio scores respectively, which demonstrates the importance of 3D information.

\vspace{0.05in}\noindent \textbf{Implementation details.} The framework is implemented in PyTorch\footnote{\url{http://pytorch.org/}}. Three networks all use Stochastic Gradient Descent (SGD) with a weight decay of 0.00005 and a momentum of 0.9. The learning rate for Temporal and Spatial network is 0.005 while 0.1 for Audio network. ReLu non-linearity \cite{nair2010rectified} and Dropout\cite{srivastava2014dropout} are applied after each fully-connected layer of the three streams.

\section{Experiments}
We evaluate our approach on four raw videos with an average length of 15 minutes. Particularly,, we divide the video into clips with 5 seconds each, and measure the average scores across clips. Since no other approach is trained and tested on our dataset, we only demonstrate ablation experiment and whole results of our approach.

\vspace{0.05in}\noindent \textbf{Metrics.} As described in \cite{gygli2016video2gif}, the commonly used metric mean Average Precision (mAP) is sensitive to video length, that is, longer video will lead to lower score. Considering that our test videos are about 3 to 5 times longer than other highlight datasets \cite{song2015tvsum,sun2014ranking}, we use the normalized version of MSD (nMSD) proposed in \cite{gygli2016video2gif} to measure the method. The nMSD refers to the relative length of the selected highlight clips at a recall rate $\alpha$, it can be defined as:
\begin{equation}
  nMSD = \frac{|G^*| - \alpha |G^{gt}|}{|V| - \alpha |G^{gt}|}, \\
\end{equation}
where $|*|$ means the length of corresponding videos. $V$  denotes the raw test video while the $G^{gt}$ and $G^*$ are the ground truth and the predicted highlights under the recall rate of $\alpha$. Note that the lower $nMSD$, the better performance is, so that the best performance occurred when $nMSD = 0$. 

\vspace{0.05in}\noindent \textbf{Results.} The results can be seen in Table \ref{tab:results}. As shown in the table, performance of the network becomes better when more information is fused into the network. It is notable that the single Audio Network has inferior performance since the audios extracted from videos are riddle with much noise, such as the voice of anchor and background music played by the player. It is concluded that our approach performs well even though no annotated video is available. More quantitative results can be seen in Figure \ref{fig:results}.

\begin{table}[t]
\begin{center}
\begin{tabular}{ccc|c|c}
\hline
temporal 	& 	spatial	&	audio	&	nMSD $\downarrow$	& mAP $\uparrow$ \\ \hline
$\surd$ &			& 			& 	15.13\% 	    & 21.51\% \\
 		&	$\surd$	& 			& 	14.41\% 	    & 21.24\% \\
		&			&	$\surd$ &   43.28\%         & 13.38\% \\
$\surd$ &	$\surd$	& 			& 	13.58\% 	    & 21.74\% \\
$\surd$ &	$\surd$	&	$\surd$ & \textbf{13.36}\% & \textbf{22.27}\% \\

\hline
\end{tabular}
\end{center}
\caption{Highlight score fusion strategies. The result achieves the best performance when three streams are fused together.}
\label{tab:results}
\end{table}

\section{Conclusion}
In this paper, we propose a multi-stream architecture to automatically generate highlights for live videos of the game 'Honor of Kings', which saves a lot of human resources. Particularly, since the edited highlight clips and the long live videos satisfy the constraint that the former gain higher scores than the latter, we take use of the existing highlight videos on 'Penguin E-sports' platform to optimize the network. On the other hand, we exploit the information from spatial, temporal and audio aspect, which further improves the performance on highlight generation. In the future, we will explore more effective techniques to make the best of inherent characteristics in game videos, namely audio information. For example, the video and audio can learn from each other via teacher-student mechanism. Besides, good performances on different categories of game videos can be easily achieved by applying transfer learning, such as transformation between the MOBA and RPG games.

{\small
\bibliographystyle{ieee_fullname}
\bibliography{egbib}
}

\end{document}